\pdfoutput=1

\documentclass[11pt,a4paper]{article}

\usepackage{ACL2023}

\usepackage{times}
\usepackage{latexsym}

\usepackage[T1]{fontenc}

\usepackage[utf8]{inputenc}

\usepackage{xspace,mfirstuc,tabulary}

\usepackage{microtype}
\usepackage{inconsolata}

\usepackage{hyperref}       
\usepackage{url}            
\usepackage{booktabs}       
\usepackage{amsfonts}       
\usepackage{nicefrac}       
\usepackage{microtype}      
\usepackage{makecell}
\usepackage{graphicx}
\usepackage{subfigure}
\usepackage{parskip}
\usepackage{float}
\usepackage{multirow}
\usepackage{array}
\usepackage{tabularx}
\usepackage{algorithm}
\usepackage{bm}
\usepackage{amsopn}
\usepackage{amsmath}
\usepackage{xcolor}
\usepackage{color}
\usepackage{wrapfig}
\usepackage{multibib}
\usepackage{booktabs}
\usepackage{verbatim}
\usepackage{colortbl}
\usepackage{bbm}
\usepackage{bbding}
\usepackage{tabularx}
\usepackage{natbib}

\definecolor{darkgreen}{RGB}{0,150,0}

\usepackage{amssymb}
\usepackage{pifont}

\newcommand{\grayline}{\rowcolor[gray]{.90}}
\newcommand{\redcell}[1]{\textcolor{red}{#1}}

\newcommand{\bluecell}[1]{\textcolor{blue}{#1}}

\usepackage[normalem]{ulem} 
\newcommand\hl{\bgroup\markoverwith
  {\textcolor{yellow}{\rule[-.5ex]{2pt}{2.5ex}}}\ULon}

\usepackage[most]{tcolorbox}

\title{Making Large Language Models Better Reasoners with Step-Aware Verifier}

\author{Yifei Li$^{\textbf{1,2}}$\Thanks{Work was done during an internship at Microsoft Research Asia.}, Zeqi Lin$^\textbf{2}$, Shizhuo Zhang$^{\textbf{2}}$, Qiang Fu$^\textbf{2}$, Bei Chen$^\textbf{2}$, \\
\textbf{Jian-Guang Lou}$^\textbf{2}$\textbf{, Weizhu Chen}$^\textbf{2}$ \\
$^1$ National Key Laboratory for Multimedia Information Processing, School of Computer Science, Peking University \\
$^2$ Microsoft Corporation \\
\small{{\texttt{\{yifeili, zeqi.lin, v-shizzhang, qifu, bei.chen, jlou, wzchen\}@microsoft.com}}} \\
\small{{\texttt{liyifei@stu.pku.edu.cn}}} \\
}

\date{}

\begin{document}
\maketitle
\begin{abstract}


Few-shot learning is a challenging task that requires language models to generalize from limited examples. Large language models like GPT-3 and PaLM have made impressive progress in this area, but they still face difficulties in reasoning tasks such as GSM8K, a benchmark for arithmetic problems. To improve their reasoning skills, previous work has proposed to guide the language model with prompts that elicit a series of reasoning steps before giving the final answer, achieving a significant improvement on GSM8K from $17.9\%$ to $58.1\%$ in problem-solving rate. In this paper, we present \textsc{DiVeRSe} (Diverse Verifier on Reasoning Step), a novel approach that further enhances the reasoning capability of language models. \textsc{DiVeRSe} has three main components: first, it generates diverse prompts to explore different reasoning paths for the same question; second, it uses a verifier to filter out incorrect answers based on a weighted voting scheme; and third, it verifies each reasoning step individually instead of the whole chain. We evaluate \textsc{DiVeRSe} on the latest language model \emph{code-davinci-002} and show that it achieves new state-of-the-art results on six of eight reasoning benchmarks (e.g., GSM8K $74.4\% \to 83.2\%$).


\end{abstract}

\section{Introduction}


Large pretrained language models (PLMs) have shown remarkable performance on various natural language processing tasks, either by few-shot learning with prompts \cite{radford2019language,le2021many,jin-etal-2022-good} or by fine-tuning \cite{houlsby2019parameter, hu2021lora, he2022towards}. However, despite the increasing size and capacity of PLMs such as GPT-3 with 175B parameters \cite{brown2020language} and PaLM with 540B parameters \cite{chowdhery2022palm}, their reasoning abilities are still limited and often require multiple steps to produce correct answers, especially for tasks involving arithmetic, commonsense, or inductive reasoning \cite{cobbe2021training}.

\begin{figure}[t]
\centering
\includegraphics[width=1.0\linewidth]{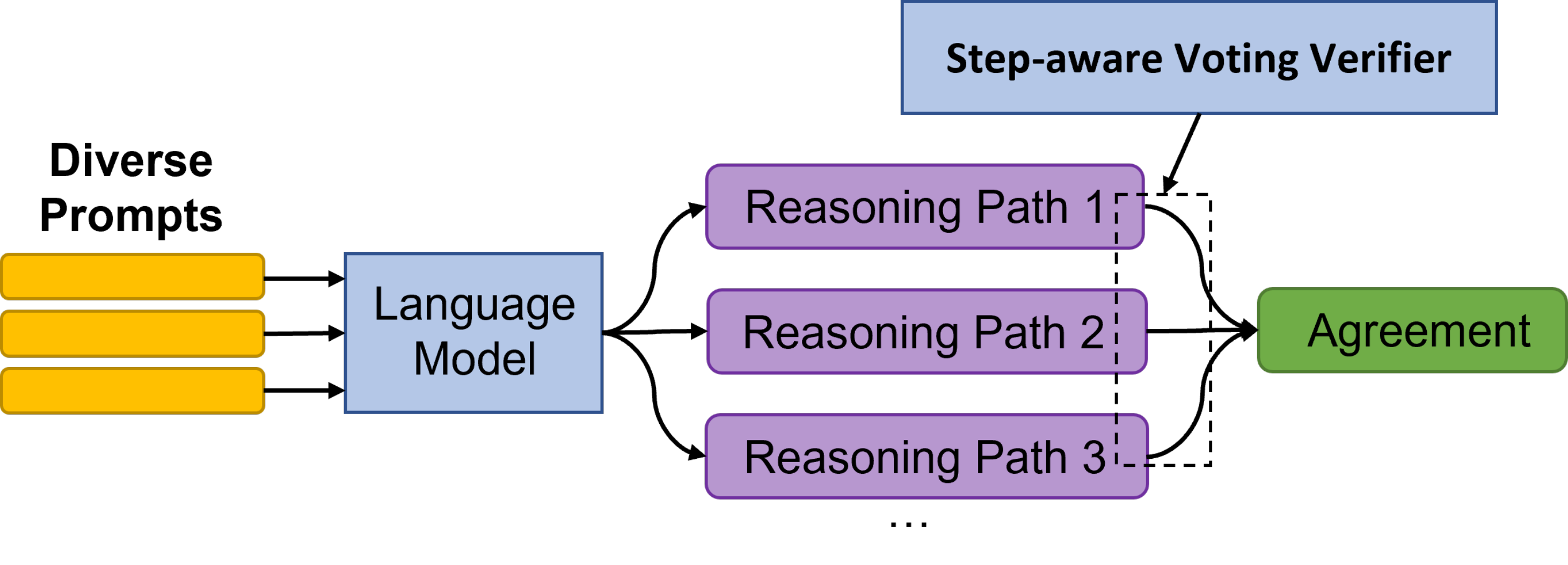}
\caption{Our proposed method, \textsc{DiVeRSe} (\textbf{Di}verse \textbf{Ve}rifier on \textbf{R}easoning \textbf{S}t\textbf{e}p).} 
\label{fig:overview}
\end{figure}

Recent works \cite{wei2022chain,least2most,fewshotreason2022,lampinen2022can} have demonstrated that PLMs possess some latent reasoning capabilities, but they need carefully designed prompts to activate them. For instance, \citet{wei2022chain} proposed chain-of-thought reasoning, which inserts multi-step reasoning paths before generating the final answers, and achieved significant improvement on the GSM8K arithmetic benchmark \cite{cobbe2021training}. \citet{selfconsistency} further introduced a voting mechanism to select the most consistent answer among different reasoning paths, and achieved state-of-the-art results on several reasoning benchmarks using the PaLM model \cite{chowdhery2022palm}. Building on these successes, this paper continues this line of research and advances the reasoning capabilities of PLMs in three aspects, as illustrated in Figure \ref{fig:overview}.

\begin{figure}[t]
\begin{tcolorbox}[colback=blue!5!white,colframe=blue!75!black,title=Chain-Of-Thought Reasoning for GSM8K Math Word Problem,fontupper=\footnotesize,fonttitle=\scriptsize]
\textbf{Q}: If there are 3 cars in the parking lot and 2 more cars arrive, how many cars are in the parking lot?

\textbf{A}:  There are 3 cars in the parking lot already. 2 more arrive. Now there are 3 + 2 = 5 cars. The answer is 5.

...

\textbf{Q}: Janet’s ducks lay 16 eggs per day. She eats three for breakfast every
morning and bakes muffins for her friends every day with four. She sells the remainder for \$2 per egg. How much does she make every day?

\textbf{A}: \textcolor{teal}{She has 16 - 3 - 4 = 9 eggs left. So she makes $2 * 9 = $18 per
day. The answer is 18.}
\end{tcolorbox}
\caption{Chain-of-thought reasoning for GSM8K math word problem. The prompt is colored black and the reasoning path produced by the language model is colored teal. This reasoning path contains two reasoning steps.}
\label{fig:cot}
\end{figure}

First, we propose to increase the diversity of reasoning paths by not only sampling from a single prompt, but also varying the prompt itself. We hypothesize that different prompts can elicit different ways of thinking, while the correct answer should be robust to these variations.
Second, we propose to use a verifier to score the quality of each reasoning path and guide the voting mechanism. We argue that not all reasoning paths are equally good or reliable, and some may contain errors or inconsistencies that can be detected by the verifier.
Third, we propose to assign a fine-grained label to each step of the reasoning path and use a step-aware verifier to attribute the correctness or wrongness of the final answer to each step. We conjecture that some steps may be correct but followed by wrong steps or vice versa, and identifying these cases can help diagnose and improve the reasoning process.


We name our method as \textsc{DiVeRSe} (diverse verifier on reasoning step) and evaluate it on eight reasoning benchmarks that require different types of reasoning skills. We use three OpenAI PLMs (\emph{davinci}, \emph{text-davinci-002}, and \emph{code-davinci-002}) and compare our results with recent state-of-the-art methods. We find that \textsc{DiVeRSe} can consistently and significantly improve the performance of PLMs on these tasks, and achieve new state-of-the-art results on six of them\footnote{Most of the previous SOTA results were achieved by self-consistency on PaLM-540B\cite{chowdhery2022palm}.}: GSM8K ($74.4\% \to 83.2\%$), AsDiv ($81.9\% \to 88.7\%$), MultiArith ($99.3\% \to 99.8\%$), SVAMP($86.6\% \to 87.0\%$), SingleEq ($79.5\% \to 94.9\%$), and CLUTRR ($67.0\%\to 95.9\%$).


Our data is publicly available at \url{https://github.com/microsoft/DiVeRSe}.

\section{Diverse Verifier on Reasoning Step}
\label{sec:consistency}

Figure \ref{fig:overview} shows the overview of \textsc{DiVeRSe}.
The key insights are three-fold:
(1) leveraging \emph{diverse prompts} to induce more diverse reasoning paths from the language models (Section \ref{section:diversity});
(2) training a \emph{voting verifier} to better derive the final answers from multiple reasoning paths (Section \ref{section:measure});
(3) leveraging \emph{step correctness} to further boost the voting verifier (Section \ref{section:granularity}).

\subsection{Diverse Prompts}
\label{section:diversity}

To reason effectively, it is beneficial to explore diverse reasoning paths, following the idea that ``\emph{All Roads lead to Rome}''.
\citet{selfconsistency} proposed to generate various reasoning paths from language models by \emph{sampling decoding}.
However, their method relies on a fixed set of exemplars for all prompts, which may introduce bias and limit the diversity of the generated reasoning paths.
To address this issue, we randomly select $M_1$ different prompts for each question, and then sample $M_2$ reasoning paths for each prompt using sampling decoding.
This way, we obtain $M=M_1\times M_2$ diverse reasoning paths for each question.\footnote{Our main experiments use $M_1=5$ and $M_2=20$.}

\subsection{Voting Verifier}
\label{section:measure}

\paragraph{Verifier.}
The verifier takes a question and a candidate reasoning path as input, and outputs the probability that the reasoning path leads to the correct answer.
We use \emph{deberta-v3-large} \cite{he2021deberta} as the backbone model, with a small scalar head that outputs predictions on the $\mathbf{[CLS]}$ token.


\paragraph{Training the verifier.}
For each training question, we generate multiple candidate reasoning paths using chain-of-thought reasoning.
We regard the reasoning paths that match the ground truth final answer as positive, and the others as negative.

\paragraph{Voting Verifier.}
\citet{selfconsistency} use \emph{majority voting} to aggregate the predictions of different reasoning paths.
This method may fail when the majority of the reasoning paths are misled, while the minority of the reasoning paths are reasonable.
We propose \emph{voting verifier}, which leverages both \emph{voting} and \emph{verifier}:

\begin{equation}
\hat{\mathbf{y}}=\underset{\mathbf{y}}{\arg\max}\sum_{i=1}^M{\mathbbm{1}_{\mathbf{y}_i=\mathbf{y}}\cdot f(\mathbf{x}_i,\mathbf{z}_i,\mathbf{y}_i)},
\end{equation}
where $\mathbbm{1}_{\mathbf{y}_i=\mathbf{y}}$ is an indicator function that returns 1 (or 0) if $\mathbf{y}_i=\mathbf{y}$ (or not), and $f(\cdot)$ is the probability produced by the verifier.

\begin{figure}[t]
\centering
\includegraphics[width=0.8\linewidth]{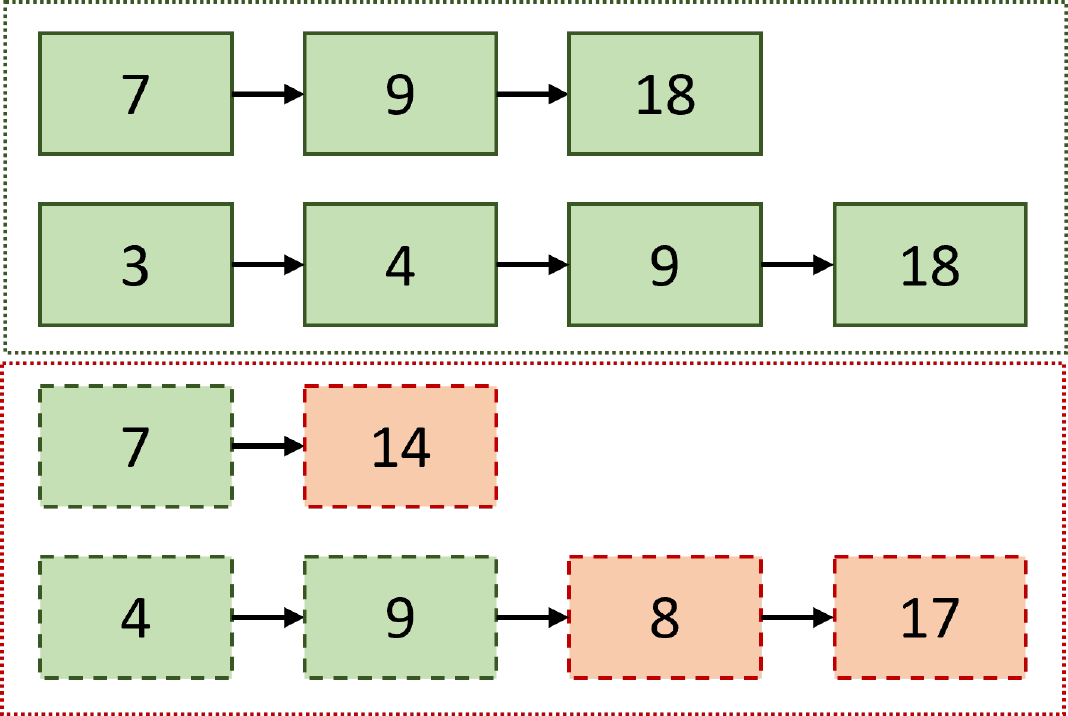}


\caption{How step-level labels are extracted. This figure shows four reasoning paths for a math word problem: the first two are positive and the bottom two are negative. The path $7\to 9\to 18$ means that the first step calculates 7, the second step calculates 9, and the third step calculates the final answer 18. For the last path, the third step (which calculates $8$) has never occurred in any positive reasoning paths, thus we regard this step and all steps after it as negative steps.}
\label{fig:step}
\end{figure}

\subsection{Step-aware Voting Verifier}
\label{section:granularity}

Each reasoning path consists of several steps.
We hypothesize that not all the steps in an incorrect reasoning path are equally wrong, and some steps may still be useful for reasoning.
To exploit this, we extend the voting verifier to a step-aware voting verifier by introducing an extended loss function:

\begin{equation}
\begin{split}
\mathcal{L} & = \mathcal{L}_0 + \alpha\cdot\mathcal{L}_1, \\
\mathcal{L}_1 = \sum_{i=1}^{|\hat{D}|}\sum_{j=1}^{|S_i|}&\text{BCE}(\text{label}_{i,j}, f'(\text{input}_i, j)).
\end{split}
\end{equation}

$\alpha$ is a hyperparameter to balance the original loss $\mathcal{L}_0$ and the step-level auxiliary loss $\mathcal{L}_1$;
$S_{i,1}, S_{i,2}, ..., S_{i,|S_i|}$ are the steps in $\mathbf{z}_i$;
$\text{label}_{i,j}$ indicates whether $S_{i,j}$ is correct or not;
$f'(\text{input}_i, j)$ represents the probability of the positive label for $S_{i,j}$.\footnote{Specifically, $f'(\text{input}_i, j)$ is predicted from the hidden state of the last token of $S_{i,j}$ in \textsc{deberta-v3-large}, similar to token classification tasks.}


\textbf{To obtain the step-level labels} (i.e., $\text{label}_{i,j}$) for negative training data with wrong answers, we design an algorithm that compares intermediate results among steps in positive/negative reasoning paths.
Figure \ref{fig:step} illustrates this algorithm.
This algorithm can not only work on math word problems, but also generalize to other reasoning tasks: we use an off-the-shelf natural language inference model, \emph{roberta-large-mnli} \cite{liu2019roberta}, to check whether two reasoning steps are semantically equivalent or not.
Given a reasoning step, if we cannot find any semantically equivalent step in the positive reasoning paths, we label it and all the subsequent steps as negative steps.


\section{Experimental Setup}

\begin{table*}[t]
\tiny
\resizebox{0.96\textwidth}{!}{
\begin{tabular}{lccccccccc}
\toprule[2pt]
Method & GSM8K & AsDiv & MultiArith & SVAMP & SingleEq & CommonsenseQA & StrategyQA & CLUTRR \\ \hline
Previous SOTA (Fine-tuning) & 57$^{a}$ & 75.3$^{b}$ & 60.5$^{c}$ & 57.4$^{d}$ & 32.5$^{e}$ & 91.2$^{f}$ & 73.9$^{g}$ & 67.0 $^{h}$ \\ 
9–12 year olds \cite{cobbe2021training} & 60 & - & - & - & - & - & - & - \\ \hline
\multicolumn{3}{l}{\underline{LaMDA 137B:}} \\
Greedy Decode & 17.1 & 49.0 & 51.8 & 38.9 & 56.6 & 57.9 & 65.4 & - \\
Self-Consistency & 27.7 & 58.2 & 75.7 & 53.3 & - & 63.1 & 67.8 & - \\ \hline
\multicolumn{3}{l}{\underline{PaLM 540B:}} \\
Greedy Decode & 56.5 & 74.0 & 94.7 & 79.0 & 79.5 & 79.0 & 75.3 & - \\
Self-Consistency & 74.4 & 81.9 & 99.3 & 86.6 & - & \textbf{80.7} & \textbf{81.6} & - \\
\bottomrule[1pt]

\multicolumn{3}{l}{\underline{GPT-3 davinci (175B):}} \\
Greedy Decode & 8.7 & 31.4 & 31.4 & 21.2 & 38.2 & 48.2 & 59.2 & 33.6 \\
Self-Consistency & 18.9 & 52.8 & 68.6 & 44.6 & 59.6 & 57.4 & 65.6 & 42.5 \\
\grayline \textbf{\textsc{DiVeRSe}} & 30.9 \redcell{(+12.0)} & 57.6 \redcell{(+4.8)} & 87.6 \redcell{(+19.0)} & 46.9 \redcell{(+2.3)} & 65.1 \redcell{(+5.5)} & 75.0 \redcell{(+17.6)} & 66.3 \redcell{(+0.7)} & 92.5 \redcell{(+50.0)} \\ \hline

\multicolumn{3}{l}{\underline{text-davinci-002:}} \\
Greedy Decode & 37.1 & 60.8 & 70.7 & 60.0 & 73.3 & 65.5 & 57.8 & 32.4 \\
Self-Consistency & 58.2 & 76.9 & 88.4 & 78.2 & 87.2 & 72.9 & 69.8 & 34.9 \\
\grayline \textbf{\textsc{DiVeRSe}} & 70.2 \redcell{(+12.0)} & 83.5 \redcell{(+6.6)} & 96.4 \redcell{(+8.0)} & 82.7 \redcell{(+4.5)} & 86.5 \bluecell{(-0.7)} & 79.2 \redcell{(+6.3)} & 74.8 \redcell{(+5.0)} & 93.8 \redcell{(+58.9)} \\ \hline

\multicolumn{3}{l}{\underline{code-davinci-002:}} \\
Greedy Decode & 55.3 & 75.5 & 88.8 & 70.5 & 87.5 & 73.4 & 72.0 & 32.9 \\
Self-Consistency & 76.7 & 86.2 & 98.6 & 85.8 & 93.7 & 77.3 & 77.6 & 35.6 \\
\grayline \textbf{\textsc{DiVeRSe}} & \textbf{82.3 \redcell{(+5.6)}} & \textbf{88.7 \redcell{(+1.5)}} & \textbf{99.8 \redcell{(+1.2)}} & \textbf{87.0 \redcell{(+1.2)}} & \textbf{94.9 \redcell{(+1.2)}} & 79.9 \redcell{(+2.6)} & 78.6 \redcell{(+1.0)} & \textbf{95.9 \redcell{(+60.3)}} \\
\bottomrule[2pt]
\end{tabular}

}
\caption{The comparison of \textsc{DiVeRSe}, \textit{Greedy Decode} and \textit{Self-Consistency}. The previous SOTA results (fine-tuned on non-gigantic pretrained transformers) are: $a$: \citet{cobbe2021training}, $b$: \citet{miao2020diverse}, $c$: \citet{roy2015solving}, $d$: \citet{pi2022reasoning}, $e$: \citet{hu2019multi}, $f$: \citet{xu2021human}, $g$: \citet{chowdhery2022palm}, $h$: \citet{sinha2019clutrr}. The parameter number of either \textit{text-davinci-002} or \textit{code-davinci-002} is hidden to us. }
\label{tab:overall_results}
\end{table*}

\subsection{Reasoning Tasks}
\paragraph{Arithmetic Reasoning.}
Following \citet{selfconsistency}, we use AsDiv \cite{miao2020diverse}, SingleEq \cite{koncel2015parsing}, MultiArith \cite{roy2015solving}, SVAMP \cite{patel2021nlp}, and GSM8K \cite{cobbe2021training}.

\paragraph{Commonsense Reasoning.}
Following \citet{selfconsistency}, we use CommonsenseQA \cite{talmor2019commonsenseqa} and StrategyQA \cite{geva2021did}.

\paragraph{Inductive Reasoning.}
We use CLUTRR \cite{sinha2019clutrr}, a diagnostic benchmark for inductive reasoning, requiring inferring kinship relations between characters in short stories.

\subsection{Details}
\paragraph{Language Models.}
We use three OpenAI language models: \emph{davinci}, \emph{text-davinci-002} and \emph{code-davinci-002}.
We use the default parameters except a temperature of $0.5$ in sampling. 

\paragraph{Exemplars.}
For arithmetic/commonsense/inductive reasoning, each prompt contains $5/7/7$ exemplars.
For \textsc{DiVeRSe}, each question has $5$ different prompts, and $20$ reasoning paths are sampled from the language model for each prompt.
For arithmetic reasoning, the exemplars are randomly sampled from the training dataset of GSM8K;
for CLUTRR, the exemplars are sampled from its training dataset, with reasoning paths synthesized by handcraft rules (detailed settings for CLUTRR are listed in Appendix \ref{sec:clutrr_detail});
for StrategyQA and CommonsenseQA, their original datasets do not contain enough exemplars with well-annotated reasoning paths, so we construct $1,000$ pseudo exemplars by ``self-teaching'' (the approach and the noise issue are discussed in Appendix \ref{sec:boosting_via_self_teaching}) from ``seed'' exemplars provided by \citet{wei2022chain}.

\paragraph{Training Datasets.}
For each task, we sample $1,000$ $\langle\text{question}, \text{answer}\rangle$ pairs from the training dataset to train the verifier.

\paragraph{Verifier.}
We fine-tune \emph{deberta-v3-large} \cite{he2021deberta} with learning rate $1\times 10^{-5}$ and batch size $128$.
For the step-aware verifier, we select the best $\alpha$ among $0.0/0.1/0.2/0.3$.

\begin{table*}[t]
\renewcommand\arraystretch{1.2}
\small
\centering
\resizebox{0.99\linewidth}{!}{
\begin{tabular}{p{0.3\textwidth}<{\raggedright}|p{0.4\textwidth}<{\raggedright}|p{0.4\textwidth}<{\raggedright}}
\toprule[2pt]
Question & Self-Consistency & \textsc{DiVeRSe} \\
\midrule[1.5pt]
\textbf{[GSM8K]} \textit{Mark decides to buy packs of old magic cards and open them to sell. He buys 3 packs for \$1800 each. He gets 1 card that is worth \$4000 and another card worth \$1000. There are 30 more cards worth an average of \$50 each. How much money profit did he make?} &

Mark bought 3 packs for \$1800 each for a total of \$1800 * 3 = 5400 \colorbox{green}{(68.1\%)}. He got 1 card worth \$4000 and another card worth \$1000 for a total of \$4000 + \$1000 = \$5000 \colorbox{green}{(72.1\%)}. There were 30 more cards worth an average of \$50 each for a total of \$50 * 30 = 1500 \colorbox{yellow}{(63.7\%)}. Mark made a profit of \$5000 + \$1500 = 6500 \colorbox{red}{(40.5\%)}. The answer is 6500.

\textbf{Answer:} \redcell{\textbf{6500 (Wrong)}}\qquad
\textbf{Score:} \colorbox{yellow}{$62.3\%$} &

The cheaper cards are worth a total of 30 * 50 = 1500 \colorbox{green}{(84.4\%)}. So all the cards together are worth 1500 + 4000 + 1000 = 6500 \colorbox{green}{(83.3\%)}. The cost for the cards was 1800 * 3 = 5400 \colorbox{green}{(80.1\%)}. So he made a profit of 6500 - 5400 = 1100 \colorbox{green}{(70.0\%)}. The answer is 1100.

\textbf{Answer:} \bluecell{\textbf{1100 (Correct)}}\qquad
\textbf{Score:} \colorbox{green}{$91.2\%$} \\
\bottomrule[2pt]
\end{tabular}
}
\caption{A GSM8K example (\textit{code-davinci-002}) with step-level scores given by the step-aware verifier. The scores can not only improve the performance but also help the understanding of where the reasoning paths start to be incorrect.}
\label{tab:step_case}
\end{table*}

\section{Main Results}
\label{section:results}

Table \ref{tab:overall_results} shows the overall experimental results.
We mainly compare \textsc{DiVeRSe} with two baselines: (1) greedily decoding a single reasoning path \cite{wei2022chain}, referred to as \emph{Greedy Decode}; (2) sampling  $100$ reasoning paths, then select the final answer via majority voting \cite{selfconsistency}, referred to as \emph{Self-Consistency}.

\subsection{Effectiveness}
Experimental results clearly demonstrate that \textsc{DiVeRSe} can bring significant and consistent improvements over recent strong baselines.
The improvements are across different models (\emph{davinci}, \emph{text-davinci-002} and \emph{code-davinci-002}) as well as different reasoning skills (eight tasks in three reasoning skills).
Taking GSM8K as an example, compared to \emph{Greedy Decoding} and \emph{Self-Consistency}, \textsc{DiVeRSe} brings improvements of $22.2\%/12.0\%$ on \emph{davinci}, $33.1\%/12.0\%$ on \emph{text-davinci-002}, and $27.0\%/5.6\%$ on \emph{code-davinci-002}.
Compared to \emph{Self-Consistency}, \textsc{DiVeRSe} achieves average improvements of $5.6\%/5.1\%/54.3\%$ on the three reasoning skills, respectively.

\subsection{Comparing to Previous SOTAs}
In Table \ref{tab:overall_results}, we also compare \textsc{DiVeRSe} with:
(1) previous SOTA results based on fine-tuning;
(2) recent SOTA results \cite{wei2022chain} based on PaLM \cite{chowdhery2022palm}, a gigantic language model with 540 billion parameters.\footnote{\textsc{DiVeRSe} can also be applied to PaLM, but PaLM is not publicly available.}

On all the five arithmetic reasoning tasks, \textsc{DiVeRSe} (with \emph{code-davinci-002}) achieves new SOTA results, with an average improvement of $6.2\%$.
On the two commonsense reasoning tasks, the performance of \textsc{DiVeRSe} is slightly lower ($-1.9\%$) than that of PaLM-based self-consistency.
We speculate that the reason might be: these two commonsense reasoning tasks are multiple-choice tasks rather than open-ended generation tasks, resulting in more false-positive exemplars in the pseudo exemplar base (Details will be discussed in Section \ref{sec:noise}).
Regarding inductive reasoning, \textsc{DiVeRSe} achieves a surprisingly good performance of $95.9\%$ on the CLUTRR task, outperforming ($+28.9\%$)  previous SOTA result with fine-tuning \cite{sinha2019clutrr}.\footnote{\citet{sinha2019clutrr} also introduced a method with $100\%$ accuracy. We do not take it into the comparison, as this method requires a domain-specific system with complicated rules to extract a knowledge graph for each input text.}

\section{Case Study}

Table \ref{tab:step_case} shows an example of step-level scores given by the step-aware verifier.
Steps in the correct reasoning path have relatively high scores, while the scores in the wrong reasoning path show where the path starts to be wrong.
This indicates that besides improving the performance, the step-aware verifier can also bring interpretability to show the step-level correctness.
We also show some extra examples of majority-voting in Table \ref{tab:code_davinci_case}.
 
\section{Analysis}
We also conduct ablation experiments and analysis to investigate the keys to the success of \textsc{DiVeRSe}.

\begin{table}[t]
\renewcommand\arraystretch{1.2}
\small
\centering
\resizebox{0.95\linewidth}{!}{
\begin{tabular}{lccc}
\toprule[2pt]
Method & GSM8K & CQA & CLUTRR \\
\midrule[1.5pt]
\multicolumn{3}{l}{\underline{davinci:}} \\
$M_1=1, M_2=100$ & 18.9 & 57.4 & 42.5 \\
\grayline $M_1=5, M_2=20$ & \textbf{21.3} & \textbf{57.5} & \textbf{45.9} \\ \hline
\multicolumn{3}{l}{\underline{text-davinci-002:}} \\
$M_1=1, M_2=100$ & 58.2 & 72.9 & 34.9 \\
\grayline $M_1=5, M_2=20$ & \textbf{61.3} & \textbf{77.3} & \textbf{35.6} \\ \hline
\multicolumn{3}{l}{\underline{code-davinci-002:}} \\
$M_1=1, M_2=100$ & 76.7 & 77.3 & 35.6 \\
\grayline $M_1=5, M_2=20$ & \textbf{80.0} & \textbf{78.8} & \textbf{43.8} \\
\bottomrule[2pt]
\end{tabular}
}
\caption{The effectiveness of diverse prompts ($\langle5, 20\rangle$) compared to pure sampling decoding \cite{selfconsistency}, under majority voting.}
\label{tab:cross_consistency}
\end{table}

\begin{table}[t]
\renewcommand\arraystretch{1.2}
\small
\centering
\resizebox{0.7\linewidth}{!}{
\begin{tabular}{lc}
\toprule[2pt]
$\langle M_1, M_2\rangle$ & \multicolumn{1}{c}{GSM8K} \\
\midrule[1.5pt]
$M_1=1, M_2=100$ & 76.7 \\
\grayline $M_1=5, M_2=20$ & \textbf{80.0} \\
$M_1=10, M_2=10$ & 79.8 \\
$M_1=100, M_2=1$ & 73.0 \\
\bottomrule[2pt]
\end{tabular}
}
\caption{GSM8K majority voting results for different $\langle M_1, M_2\rangle$ settings on \emph{code-davinci-002}.}
\label{tab:mix}
\end{table}

\subsection{The Effectiveness of Diverse Prompts}
\label{sec:power_of_diverse_prompts}

\begin{figure}[t]
\centering
\subfigure{
\includegraphics[width=3.6cm]{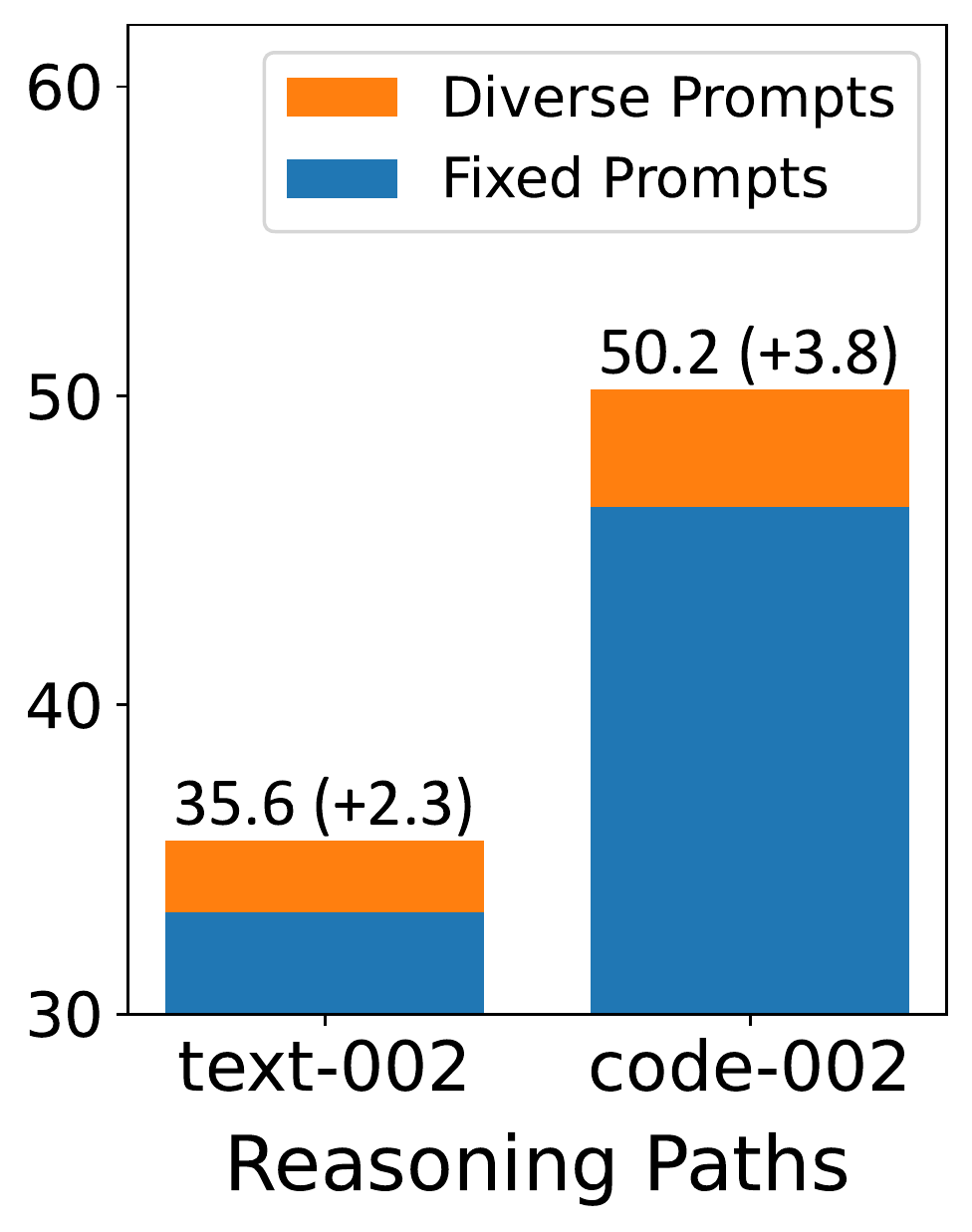}
\label{fig:num_of_different_rps}
}
\subfigure{
\includegraphics[width=3.6cm]{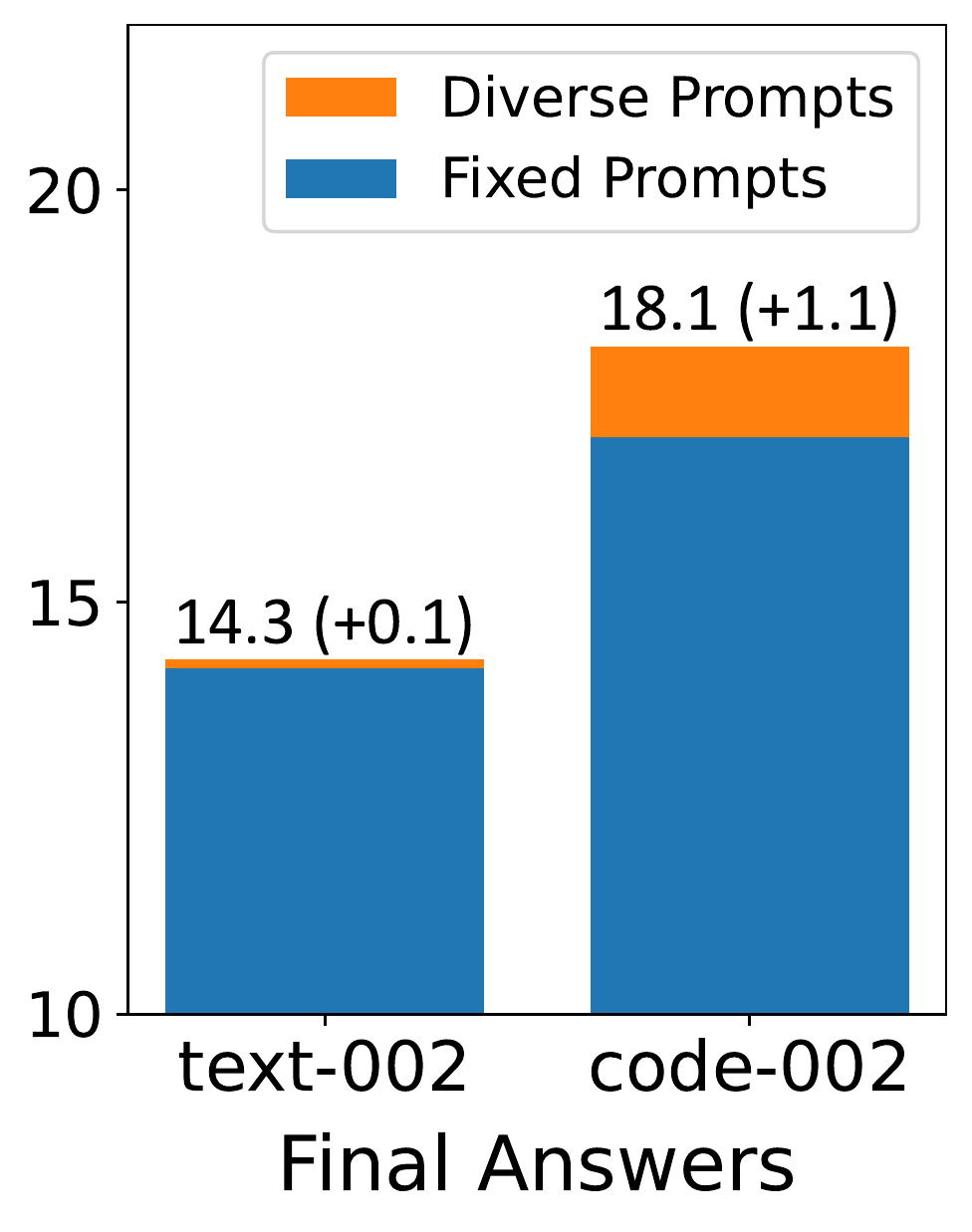}
\label{fig:num_of_different_final_answers}
}
\caption{Diverse prompts increase the diversity of GSM8K reasoning paths and their final answers. This is beneficial for the voting verifier. Left: the average number of distinct reasoning paths per question (we consider two reasoning paths to be the same if they have the same intermediate result chain as shown in Figure \ref{fig:step}). Right: the average number of distinct final answers per question.}
\label{fig:diversity}
\end{figure}

By diversifying both prompts and reasoning paths ($\langle M_1=5, M_2=20\rangle$), we consistently improve performance over the sampling decoding approach ($\langle M_1=1, M_2=100\rangle$) of \citet{selfconsistency}, as shown in Table \ref{tab:cross_consistency}. Both methods use majority voting. Table \ref{tab:mix} further reveals that neither only using diverse prompts nor only using sampling is optimal. In other words, \emph{the best performance is achieved by combining diverse prompts and sampling}. Moreover, Figure \ref{fig:diversity} demonstrates that \textit{diverse prompts lead to more diverse reasoning paths}. We hypothesize that this diversity contributes to the performance improvement by: (1) making correct results more distinguishable from varied errors during inference; and (2) providing more diverse negative samples for enhancing the verifier's generalizability during training.


\subsection{The Effectiveness of Voting Verifier}
\label{sec:power_of_voting_verifier}

We compare three algorithms to conclude the agreement from diverse reasoning paths: majority voting, verifier, and voting verifier.
Table \ref{tab:the_power_of_expectation} shows the results. 
\emph{Compared to majority voting, our voting verifier can significantly and consistently boost reasoning performance across different tasks and different language models}.
Verifier without voting often outperforms majority voting, but extending it to voting verifier can further boost the performance.




\begin{table}[t]
\renewcommand\arraystretch{1.2}
\small
\centering
\resizebox{0.99\linewidth}{!}{
\begin{tabular}{lccc}
\toprule[2pt]
Method & \multicolumn{1}{c}{GSM8K} & \multicolumn{1}{c}{CQA} & \multicolumn{1}{c}{CLUTRR} \\
\midrule[1.5pt]
\multicolumn{3}{l}{\underline{davinci:}} \\
Voting & 21.3 & 57.4 & 45.9 \\
Verifier & 27.0 & 74.1 & \textbf{93.2} \\
\grayline Voting Verifier & \textbf{30.6} & \textbf{75.0} & 92.5 \\ \hline
\multicolumn{3}{l}{\underline{text-davinci-002:}} \\
Voting & 61.3 & 77.3 & 35.6 \\
Verifier & 62.7 & 77.9 & \textbf{93.8} \\
\grayline Voting Verifier & \textbf{68.9} & \textbf{79.2} & \textbf{93.8} \\ \hline
\multicolumn{3}{l}{\underline{code-davinci-002:}} \\
Voting & 80.0 & 75.4 & 43.8 \\
Verifier & 65.9 & \textbf{78.8} & \textbf{95.9} \\
\grayline Voting Verifier & \textbf{82.3} & \textbf{78.8} & \textbf{95.9} \\
\bottomrule[2pt]
\end{tabular}
}
\caption{The effectiveness of voting verifier. All exepriments in this table use $\langle M_1,M_2 \rangle=\langle5, 20\rangle$.}
\label{tab:the_power_of_expectation}
\end{table}

\begin{figure}[t]
\centering
\subfigure[The number of correct reasoning paths containing redundant steps.]{
\includegraphics[width=3.5cm]{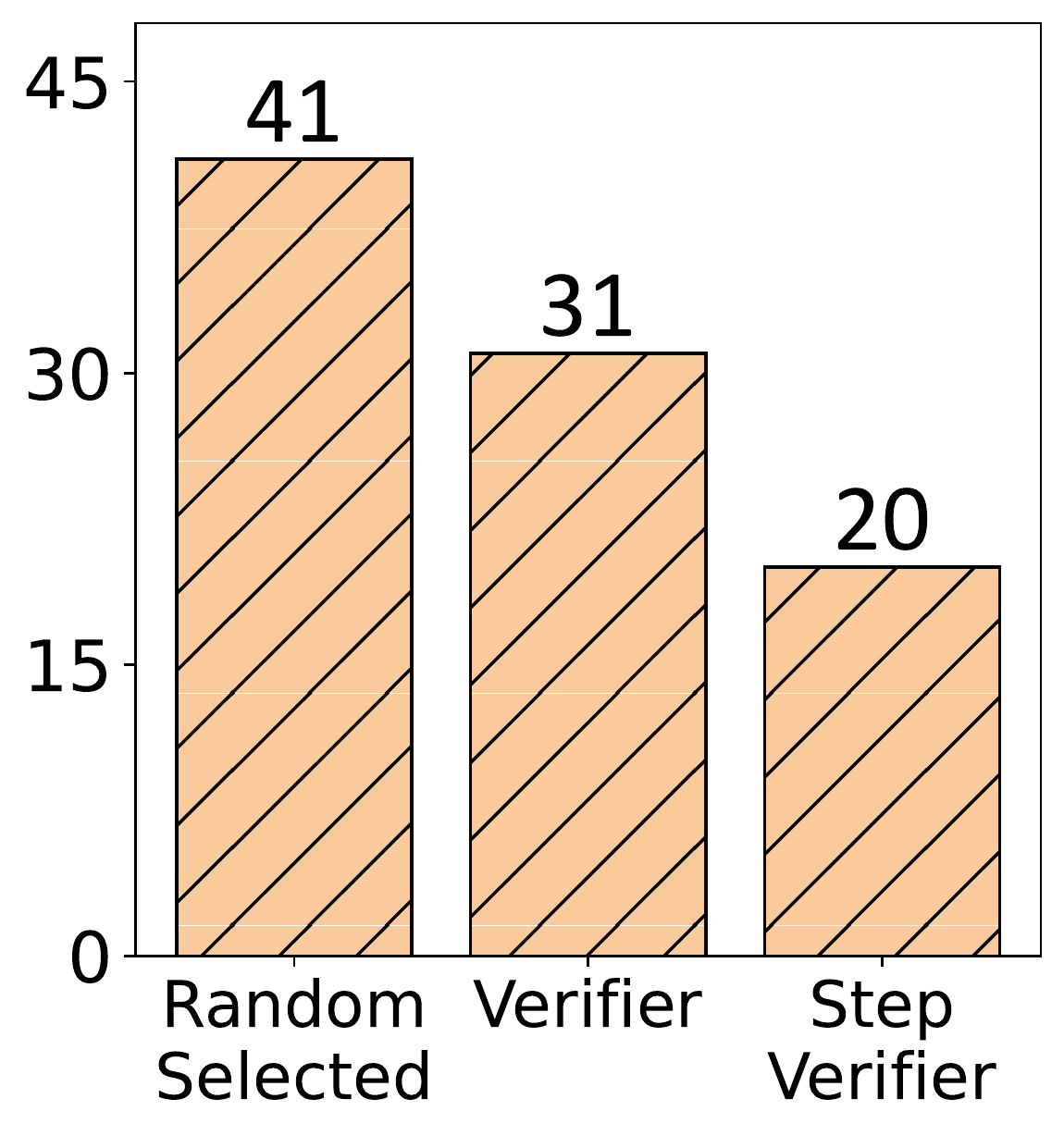}
\label{fig:step_correct_analysis}
}
\subfigure[With the step-aware mechanism, incorrect paths contain more correct steps.]{
\includegraphics[width=3.5cm]{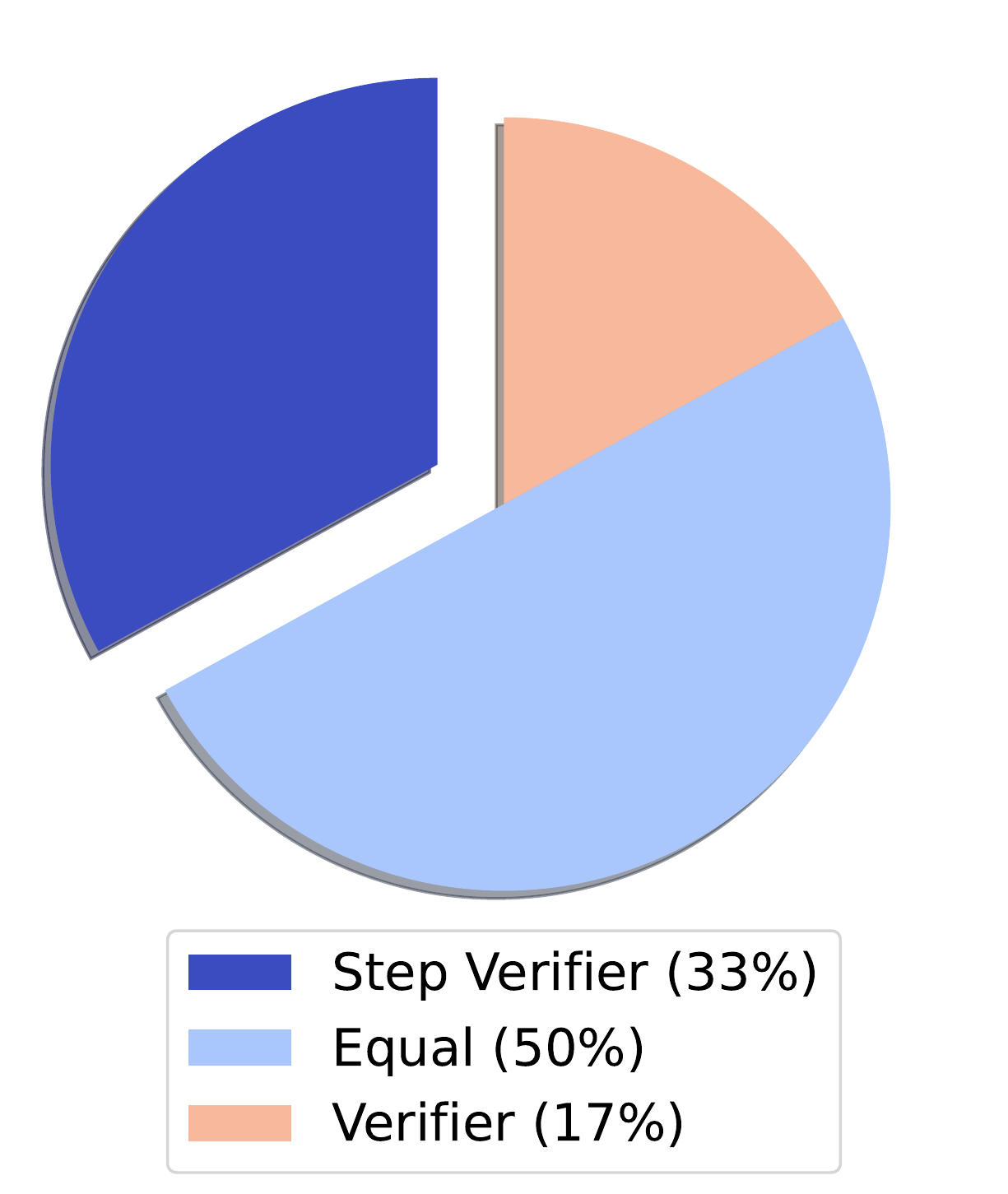}
\label{fig:step_incorrect_analysis}
}
\caption{Human evaluation on GSM8K shows the effectiveness of the step-aware mechanism for verifier.}
\end{figure}


\subsection{The Effectiveness of Step-aware Verifier}
\label{sec:step_level}


We evaluate the impact of incorporating step-level information into the voting verifier of \textsc{DiVeRSe}.
Table \ref{tab:step_level} shows the performance of \textsc{DiVeRSe} with and without the step-aware mechanism on both the GSM8K  and the CommonsenseQA datasets.
We find that \emph{using the step-aware verifier improves the performance in most of the experiments}.
The only exception is \emph{code-davinci-002} on GSM8K, where the step-aware verifier slightly lowers the performance.
We hypothesize that \emph{code-davinci-002} is more capable of generating high-quality reasoning paths, and thus does not benefit much from the step-level information.

\paragraph{Detailed Human Evaluation of Reasoning Steps.}
We further analyze the quality of generated reasoning steps, by asking human annotators to judge whether the GSM8K reasoning steps produced by \textsc{DiVeRSe} (with/without step-aware mechanism) are good or not.
Here ``good'' means not only correct formulas and calculation results but also textual fluency and logical coherence.


We further examine the quality of the reasoning steps generated by \textsc{DiVeRSe} (with/without step-aware mechanism) for GSM8K, by asking human annotators to rate them based on correctness, fluency, and coherence.
For each test question, we compare three reasoning paths produced by \textit{code-davinci-002}: the one with the highest verifier score, the one with the highest step-aware verifier score, and a randomly chosen one.
The annotators (master students) label any incorrect or unsatisfactory reasoning steps in each path (single-blind) and explain why.
We collect annotations for 200 test questions, half of which have correct final answers from all three paths, and half of which have incorrect final answers from all three paths.

We find that \textbf{all the reasoning paths with correct final answers are also correct in every intermediate step}, which shows that \textit{code-davinci-002} can reliably generate accurate reasoning steps, not just lucky guesses.
However, we also find that \textbf{many of the correct reasoning paths have unnecessary steps}.
Figure \ref{fig:step_correct_analysis} shows that $40\%$ of the random paths have redundant steps, and the verifier can lower this percentage to $31\%$.
We also find that \textbf{the step-aware verifier can further eliminate redundant reasoning steps} from $31\%$ to $20\%$.


\begin{figure}[t]
	\centering
	\includegraphics[width=0.98\columnwidth]{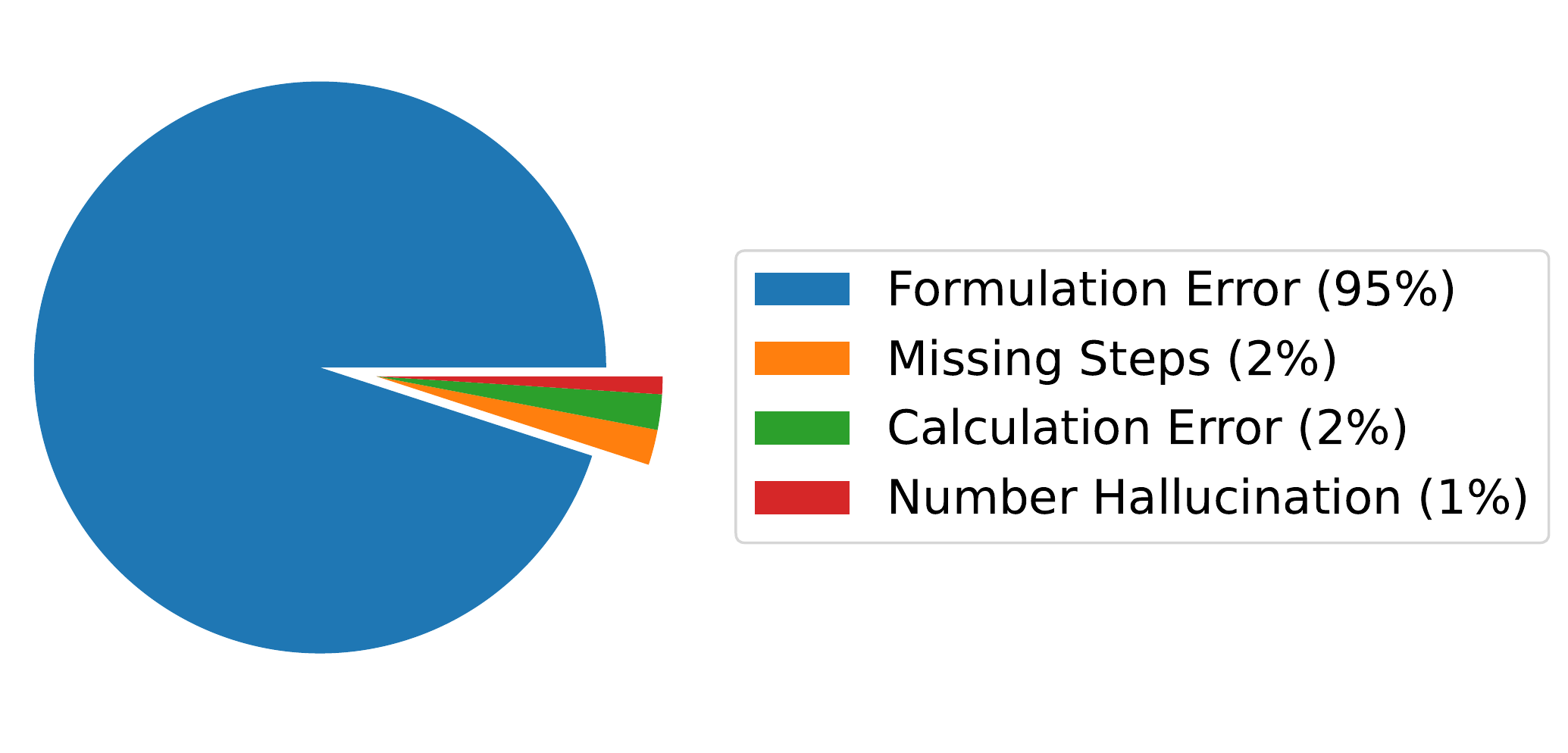}
	\caption{The distribution of error types in incorrect reasoning steps.}
    \label{fig:step_error_type_analysis}
\end{figure}


Furthermore,  for the incorrect reasoning paths, we find that \textbf{the step-aware mechanism helps produce more correct steps before making mistakes}.
For each failed test question, we compare the number of correct steps in the path with the highest verifier score and the path with the highest step-aware verifier score (by human evaluation).
Figure \ref{fig:step_incorrect_analysis} shows that for $33\%$/$17\%$ of the failed test cases, the step-aware verifier generates more/fewer correct steps than the verifier without the step-aware mechanism.


\begin{table}[t]
\renewcommand\arraystretch{1.2}
\small
\centering
\resizebox{0.99\linewidth}{!}{
\begin{tabular}{lcc}
\toprule[2pt]
 & \multicolumn{1}{c}{GSM8K} & \multicolumn{1}{c}{CommonsenseQA} \\
\midrule[1.5pt]
\multicolumn{3}{l}{\underline{davinci:}} \\
\textsc{DiVeRSe} (without step) & 30.6 & 75.0 \\
\grayline \textsc{DiVeRSe} (with step) & \textbf{30.9} & \textbf{76.0} \\ \hline
\multicolumn{3}{l}{\underline{text-davinci-002:}} \\
\textsc{DiVeRSe} (without step) & 68.9 & 79.2 \\
\grayline \textsc{DiVeRSe} (with step) & \textbf{70.2} & \textbf{79.8} \\ \hline
\multicolumn{3}{l}{\underline{code-davinci-002:}} \\
\textsc{DiVeRSe} (without step) & \textbf{82.3} & 78.8 \\
\grayline \textsc{DiVeRSe} (with step) & 81.5 & \textbf{79.9} \\ \hline
\bottomrule[2pt]
\end{tabular}
}
\caption{The effectiveness of step-aware voting verifier, with $\langle M_1,M_2 \rangle=\langle5, 20\rangle$.}
\label{tab:step_level}
\end{table}


\paragraph{Step Error Types.} Figure \ref{fig:step_error_type_analysis} shows the distribution of error types in the incorrect reasoning steps.
We see that $95\%$ of the errors are caused by incorrect formulations (i.e., using wrong intermediate results or operators and generating invalid formulas, which lead to incorrect answers).
We also see that, although \textit{code-davinci-002} often makes division calculation errors (e.g., $10/3=3$), both the verifier and the step-aware verifier can effectively assign low scores to such paths, thus improving the performance.


\subsection{How Many Diverse Outputs Do We Need?}

\begin{figure}[t]
	\centering
	\hspace*{-0.8cm}
	\includegraphics[width=0.55\textwidth]{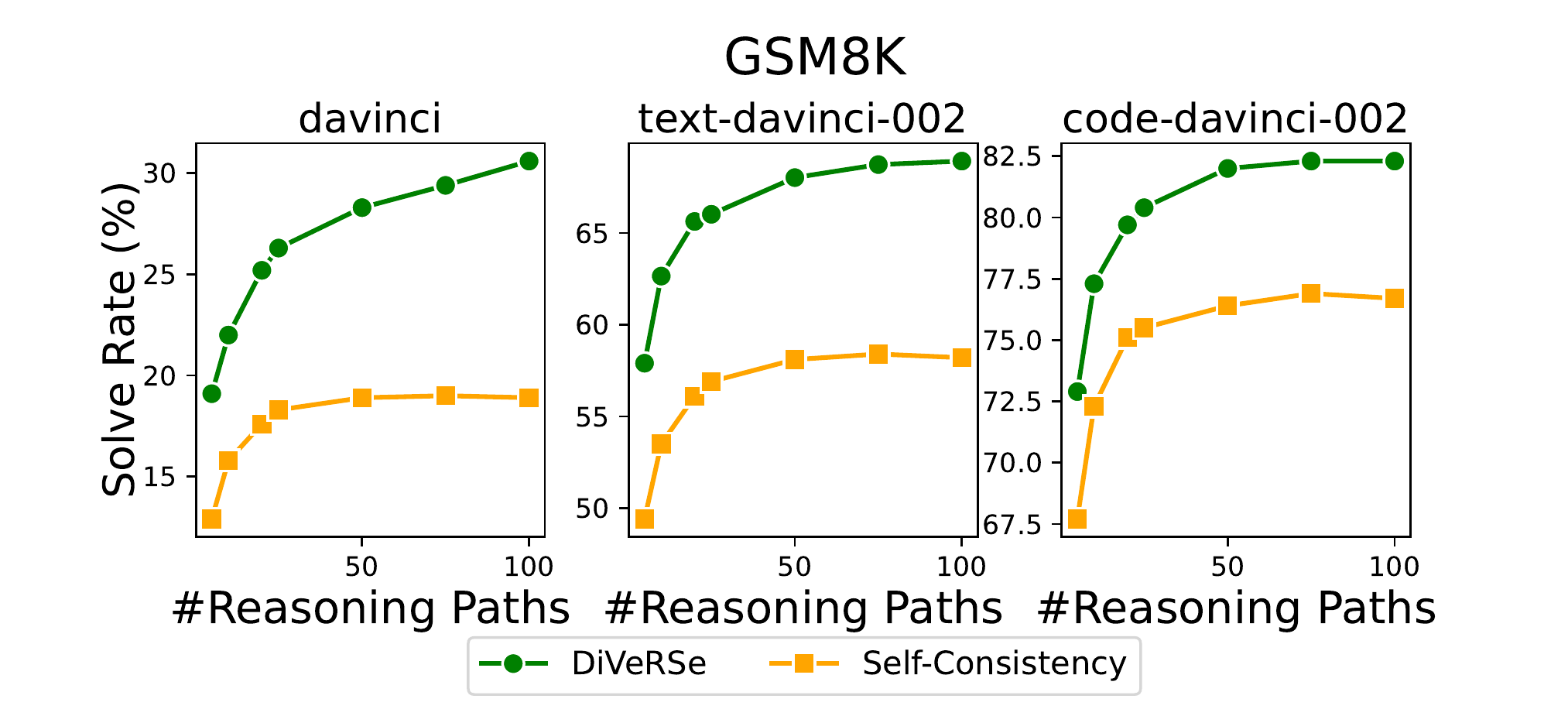}
	\caption{GSM8K accuracy at different $M$ values (how many reasoning paths are used for each question).}
	\label{fig:diverse_outputs_number}
\end{figure}

Figure \ref{fig:diverse_outputs_number} shows the accuracy at different $M$ values, where $M$ is the number of reasoning paths sampled from the $100$ generated paths for each question.
We observe that:
(1) the accuracy increases with more reasoning paths, but the improvement becomes marginal at $M\geq50$;
(2) \textsc{DiVeRSe} outperforms self-consistency significantly and consistently at different $M$ values.

\subsection{How Many Training Data Do We Need?}

\begin{figure}[t]
	\centering
	\includegraphics[width=0.4\textwidth]{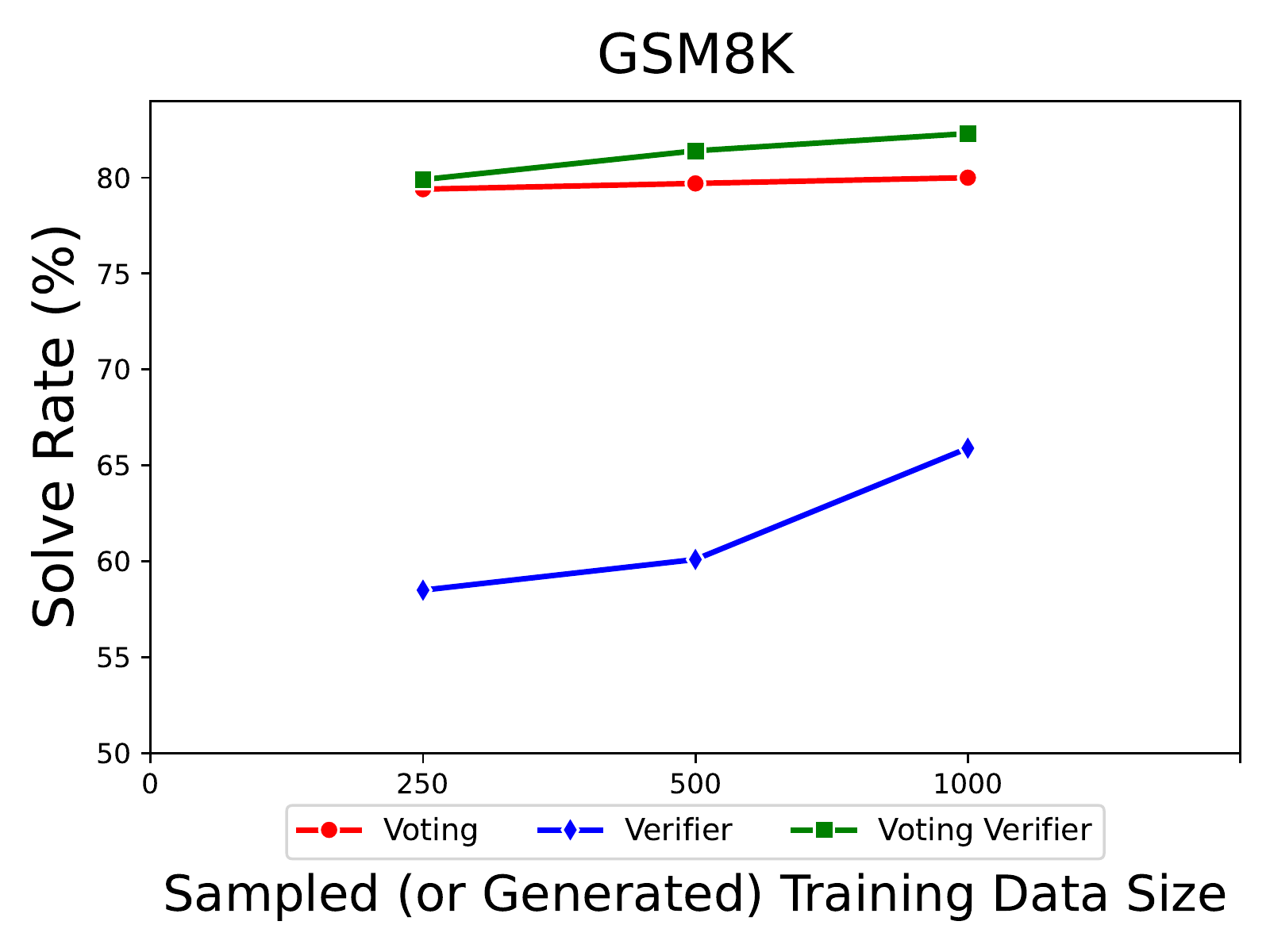}
	\caption{\textsc{DiVeRSe} performance (\emph{code-davinci-002}) on GSM8K with different sizes of the training dataset (without labeled reasoning paths).}
	\label{fig:number_of_exemplars}
\end{figure}

\textsc{DiVeRSe} requires a dataset with reasoning paths for training the verifier.
Figure \ref{fig:number_of_exemplars} shows how the size of this dataset affects the performance.
We observe that:
the performance is only reduced by about $2\%$, even if the size of training data is cut by $75\%$ (from $1,000$ to $250$).
With the same reasoning paths, voting verifier performs better than majority voting, while verifier without voting causes significant performance drops.

\subsection{The Impact of the Number of Exemplars}
\begin{figure}[t]
	\centering
	\includegraphics[width=0.4\textwidth]{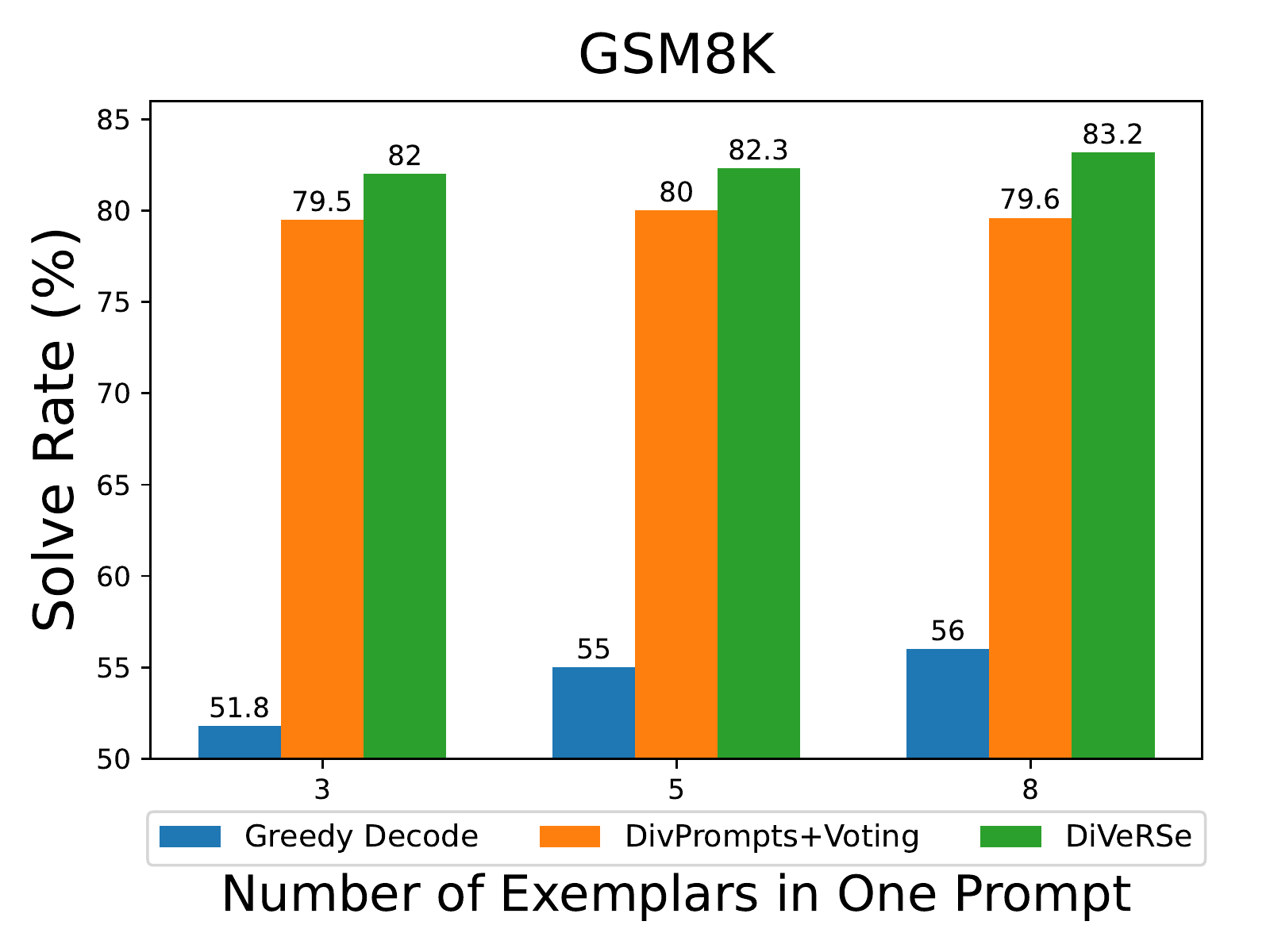}
	\caption{\textsc{DiVeRSe} performance (\emph{code-davinci-002}) on GSM8K when each prompt contains $3/5/8$ exemplars.}
	\label{fig:number_of_in_context_demos}
\end{figure}

We conduct experiments for $k=3/5/8$ ($k$ is the number of exemplars used in each prompt) on GSM8K.
Figure \ref{fig:number_of_in_context_demos} shows the results.
We observe that: \emph{using 8 exemplars in each prompt can further boost the accuracy of GSM8K to $83.2\%$.}

\section{Related Work}

\paragraph{Reasoning Skills.}
Researchers in the literature have proposed many benchmarks requiring various reasoning skills, including
commonsense reasoning \cite{zellers-etal-2018-swag,talmor2019commonsenseqa,https://doi.org/10.48550/arxiv.1908.05739,geva2021did}
numerical reasoning \cite{dua-etal-2019-drop}, multi-hop reasoning \cite{yang-etal-2018-hotpotqa}, arithmetic reasoning \cite{koncel2015parsing,roy2015solving,miao2020diverse,patel2021nlp,cobbe2021training}, logical reasoning \cite{https://doi.org/10.48550/arxiv.2007.08124,https://doi.org/10.48550/arxiv.2002.04326}, inductive reasoning \cite{sinha2019clutrr} and tabular reasoning \cite{chen-etal-2020-hybridqa,https://doi.org/10.48550/arxiv.2105.07624}.

\paragraph{Reasoning with Symbolic Systems.}
Much research in the literature enhances the reasoning capabilities of machine learning systems by exploiting symbolic systems,
including knowledge graphs \cite{mihaylov-frank-2018-knowledgeable,bauer-etal-2018-commonsense,kundu-etal-2019-exploiting,10.1609/aaai.v33i01.33017208,lin-etal-2019-kagnet,ding-etal-2019-cognitive,https://doi.org/10.48550/arxiv.2005.00646,wang2022multi-level}, or question taxonomies \cite{dua-etal-2019-drop,andor-etal-2019-giving,hu-etal-2019-multi,wang-etal-2022-logic}.
Although these methods work well on specific benchmarks, they usually require domain-specific designs and human efforts, thus limiting the generalizability.

\paragraph{Reasoning via Language Models.}
This line of work aims to address reasoning tasks in a general sequence-to-sequence manner, empowered by reasoning-aware pre-training or fine-tuning of language models.
For example, \citet{deng-etal-2021-reasonbert} proposed to train the language model with crawled data from the internet;
\citet{asai-hajishirzi-2020-logic} proposed a logic-guided data augmentation method to pre-train the language model; 
\citet{shen2021generate,cobbe2021training} proposed to train a verifier to rank solutions sampled from fine-tuned language models;
\citet{geva-etal-2020-injecting,yoran-etal-2022-turning,campagna-etal-2020-zero,wang-etal-2022-logic} proposed to equip language models with reasoning abilities by generating training examples with human-designed templates;
\citet{pi2022reasoning} proposed to inject reasoning capabilities into language models by continual pre-training on program execution data.

\paragraph{Reasoning via Prompting Gigantic Language Models.}
Gigantic language models like GPT-3 \cite{brown2020language} have demonstrated impressive few-shot learning capabilities in many tasks and have attracted many research interests on making gigantic language models better few-shot learners \cite{https://doi.org/10.48550/arxiv.2102.09690,https://doi.org/10.48550/arxiv.2104.08315,https://doi.org/10.48550/arxiv.2110.15943,liu-etal-2022-makes,https://doi.org/10.48550/arxiv.2104.08786,https://doi.org/10.48550/arxiv.2112.08633,https://doi.org/10.48550/arxiv.2202.12837}.
However, these methods struggle to address tasks requiring reasoning skills.
To mitigate this, recently there is a line of research that focuses on unleashing the reasoning capabilities of gigantic language models via better prompting strategies.
\citet{wei2022chain} proposed \emph{chain-of-thought reasoning}, of which the key insight is the insertion of multi-step reasoning paths before generating the final answers;
\citet{selfconsistency} proposed to improve chain-of-thought reasoning via \emph{self-consistency}, of which the key insight is to conclude the most consistent answer from different reasoning paths sampled from the language model;
\citet{least2most,https://doi.org/10.48550/arxiv.2205.09712} proposed to leverage gigantic language models to decompose questions into sub-questions, thereby addressing them in an iterative manner;
\citet{fewshotreason2022} proposed that gigantic language models can even be good zero-shot reasoners, by designing prompts that can induce language models to do reasoning step-by-step;
\citet{lampinen2022can} proposed building a prompt by selecting examples and explanations together, thus substantially improving performance over selecting examples alone.
Despite their great successes, these works come with their limitations.
This paper is a continuation of this line of research, focusing on diverse verifier on reasoning steps.

\section{Conclusion and Future Work}

In this paper, we present \textsc{DiVeRSe}, a novel and general method to enhance the reasoning abilities of large language models. Our method builds on the idea of prompting language models with multi-step reasoning paths, but introduces three key innovations: diverse prompts, voting verifier, and stepwise verifier. The latter is especially novel and effective, as it verifies each reasoning step separately and we provides a detailed analysis of the model's behavior in each step. We demonstrate the superiority of \textsc{DiVeRSe} through extensive experiments. For instance, using \emph{code-davinci-002}, our method achieves state-of-the-art performance on most reasoning tasks, surpassing the 540B PaLM model with previous prompting techniques.

There are many directions for our future work.
(1) As discussed in Appendix \ref{sec:noise}, we will continue to investigate how to reduce or recognize false positive pseudo exemplars.
(2) We plan to investigate mechanisms to produce better diverse prompts than simple sampling.
(3) We will extend \textsc{DiVeRSe} to other tasks and continue to design better prompting techniques to elicit the power of gigantic language models. 

\section{Limitations}

\paragraph{Computing Resources.} Despite the surprising performance it achieves, our framework needs to be applied to large language models like GPT-3 or PaLM. Inference with these models costs more time and budgets than fine-tuning models like RoBERTa \cite{liu2019roberta}.

\paragraph{Faithfulness.} Although \textsc{DiVeRSe} can significantly improve the accuracy of final answers, we still cannot guarantee that the reasoning paths produced by the language models are 100 percent faithful.
This is the key challenge and future direction for this line of research (chain-of-thought reasoning).

\paragraph{More Training Data.} \textsc{DiVeRSe} needs more labeled data with well-annotated reasoning paths to construct diverse prompts, and it also needs a training dataset for supervising the verifier.
However, from another point of view, this limitation can also be regarded as a contribution that studies how chain-of-thought reasoning can be further improved if we have more training data than just a few exemplars.

\paragraph{Human Evaluation of Reasoning Steps.} We use human evaluation to measure the quality of the intermediate steps in reasoning paths since few current works provide reliable frameworks to evaluate the quality of reasoning steps.

\bibliography{anthology,custom}
\bibliographystyle{acl_natbib}

\clearpage

\appendix
This is the Appendix for the paper: ``On the Advance of Making Language Models Better Reasoners''.


\section{Preliminaries}

\paragraph{Prompting.}

Prompting means prepending a few exemplars to the task input $\mathbf{x}$ and
generating the output $\mathbf{y}$ from the pretrained language model:
\begin{equation}
    p(\mathbf{y}|C,\mathbf{x}) = \prod_{t=1}^{|\mathbf{y}|} p_{\text{LM}}(y_t | C, \mathbf{x}, y_{<t}),
\end{equation}
where $C$ is the concatenation of $K$ exemplars:
\begin{equation}
C = (\overline{\mathbf{x}}_1,\overline{\mathbf{y}}_1);(\overline{\mathbf{x}}_2,\overline{\mathbf{y}}_2);...;(\overline{\mathbf{x}}_K,\overline{\mathbf{y}}_K).
\end{equation}
We denote \textbf{prompt} as the concatenation of the exemplars $C$ and the input $\mathbf{x}$.

\paragraph{Reasoning Paths.}

For reasoning tasks that aim to generate an answer $\mathbf{y}$ for a question $\mathbf{x}$, \citet{wei2022chain} proposed the insertion of a reasoning path $\mathbf{z}$ before generating the answer $\mathbf{y}$:
\begin{equation}
C'=(\overline{\mathbf{x}}_1,\overline{\mathbf{z}}_1,\overline{\mathbf{y}}_1);...;(\overline{\mathbf{x}}_K,\overline{\mathbf{z}}_K,\overline{\mathbf{y}}_K),
\end{equation}
where $\mathbf{z}_i$ is a text \textbf{reasoning path} of how the answer $\mathbf{y}_i$ is reasoned step-by-step for question $\mathbf{x}_i$.

Then, during inference, a reasoning path $\mathbf{z}$ will be generated before the answer $\mathbf{y}$:
\begin{equation}
    p(\mathbf{y}|C',\mathbf{x}) = p(\mathbf{z}|C',\mathbf{x})\cdot p(\mathbf{y}|C',\mathbf{x}, \mathbf{z}).
\end{equation}

Figure \ref{fig:background} demonstrates this idea in arithmetic reasoning (GSM8K), and Table \ref{tab:dataset_exemplars} demonstrates this idea in commonsense reasoning (StrategyQA) and inductive reasoning (CLUTRR).

\begin{figure}[t]
\begin{tcolorbox}[colback=blue!5!white,colframe=blue!75!black,title=Prompt \& Samples,fontupper=\footnotesize,fonttitle=\scriptsize]
\textbf{Q}: If there are 3 cars in the parking lot and 2 more cars arrive, how many cars are in the parking lot?

\textbf{A}:  There are 3 cars in the parking lot already. 2 more arrive. Now there are 3 + 2 = 5 cars. The answer is 5.

...

\textbf{Q}: Janet’s ducks lay 16 eggs per day. She eats three for breakfast every
morning and bakes muffins for her friends every day with four. She sells the remainder for \$2 per egg. How much does she make every day?

\textbf{A}:
\tcbline
\textcolor{gray}{[Sample 1]}
She has 16 - 3 - 4 = 9 eggs left. So she makes $2 * 9 = $18 per
day. The answer is 18. (\checkmark)
\tcbline
\textcolor{gray}{[Sample 2]} This means she uses 3 + 4 = 7 eggs every day. So in total she sells 7 * $2 = $14 per day. The answer is 14. (\XSolidBrush)
\tcbline
\textcolor{gray}{[Sample 3]} She eats 3 for breakfast, so she has 16 - 3 = 13 left. Then she bakes muffins, so 13 - 4 = 9 eggs left. She makes 9 * \$2 = \$18. The answer is 18. (\checkmark)
\end{tcolorbox}
\caption{Prompting large language models to generate different reasoning paths, then selecting the final answer via majority voting \cite{selfconsistency}.}
\label{fig:background}
\end{figure}

\begin{table}[t]
\begin{tabularx}{\columnwidth} { 
    >{\raggedright\arraybackslash}X 
}
\toprule[2pt]
{\small \textbf{\textsc{[StrategyQA]}} Yes or no: Could a llama birth twice during War in Vietnam
(1945-46)? $\triangleright$ \emph{The War in Vietnam was 6 months. The gestation period for a llama is 11 months. So a llama could not give birth twice
during the War in Vietnam. The answer is \textbf{\underline{no}}.}} \\ \hline
{\small \textbf{\textsc{[CLUTRR]}}  Roy was eating lunch with his son John and his wife Mary. What kind of relative is John to Mary? $\triangleright$ \emph{John is the son of Roy. Roy is the husband of Mary. Thus, John is the son of Mary. The answer is \textbf{\underline{son}}.}} \\
\bottomrule[2pt]
\caption{Besides arithmetic reasoning, we also investigate commonsense and inductive reasoning.}
\label{tab:dataset_exemplars}
\end{tabularx}
\vspace{-10mm}
\end{table}

\begin{table*}[t]
\renewcommand\arraystretch{1.2}
\centering
\begin{tabular}{lcp{0.7\textwidth}<{\raggedright}}
\toprule[2pt]
Dataset & $N$ & Example Question \\
\midrule[1.5pt]
GSM8K & 1319 & James decides to run 3 sprints 3 times a week.  He runs 60 meters each sprint.  How many total meters does he run a week? \\ \hline
AsDiv & 2096 & Seven red apples and two green apples are in the basket. How many apples are in the basket? \\ \hline
MultiArith & 600 & The school cafeteria ordered 42 red apples and 7 green apples for students lunches. But, if only 9 students wanted fruit, how many extra did the cafeteria
end up with? \\ \hline
SVAMP & 1000 & Paco had 26 salty cookies and 17 sweet cookies. He ate 14 sweet cookies and 9 salty cookies. How many salty cookies did Paco have left? \\ \hline
SingleEq & 508 & Terez has 44 cows on his farm. 50 percent of the cows are female, and 50 percent of the females are pregnant. How many pregnant female cows does Terez have? \\ \hline
CommonsenseQA & 3387 & Sammy wanted to go to where the people were. Where might he go? Options: (a) race track (b) populated areas (c) desert (d) apartment (e) roadblock \\ \hline
StrategyQA & 2280 & Could you go to New York Public Library and the Six Flags Great Escape in the same day? \\ \hline
CLUTRR & 447 & Kelly and her mother Ernest made breakfast together. Constance and her husband Ernest wanted a child badly What kind of relative is Kelly to Constance? The possible relationships are: sister, son, aunt, granddaughter, father, grandfather, grandmother, mother-in-law, uncle, niece, mother, brother, daughter, nephew, grandson, son-in-law, father-in-law, daughter-in-law. \\
\bottomrule[2pt]
\end{tabular}
\caption{Reasoning benchmarks we use in this paper with examples. $N$ means the number of test cases.}
\label{tab:data_stat}
\end{table*}

\section{Boosting Reasoning Paths via Self-Teaching}
\label{sec:boosting_via_self_teaching}

In this section, we first introduce self-teaching, the method we use to construct a larger exemplar base when the original dataset does not contain enough data with well-annotated reasoning paths (Appendix \ref{sec:self_teaching}). We then discuss the noise issue when facing multiple-choice tasks (Appendix \ref{sec:noise}).

\subsection{Self Teaching}
\label{sec:self_teaching}
A critical issue of \textsc{DiVeRSe} is \textbf{how to provide diverse prompts}.\footnote{\citet{selfconsistency} tried an ensemble-based approach, i.e., to permutate exemplars in the original prompt.
However, this strategy does not increase diversity in terms of exemplars.}
Supposing that there is an exemplar base $E$, we can sample $K$ exemplars from it to construct a prompt, and repeat this $M_1$ times independently to construct $M_1$ prompts with diverse exemplars.

For scenarios that do not have sufficient exemplars (i.e., $|E| < K * M_1$), we propose to \textbf{bootstrap the diversity of prompts by ``self-teaching''}, i.e., generating pseudo reasoning paths from a few exemplars and some $\langle\text{question}, \text{answer}\rangle$ pairs without reasoning paths.\footnote{This is motivated by \citet{zelikman2022star}.}
Suppose that $D$ is a dataset without reasoning paths, consisting of $(\mathbf{x}, \mathbf{y}^*)$ pairs.
Given the small exemplar base $E$, for each $(\mathbf{x,y^*})\in D$, we can use prompting to generate a reasoning path $\mathbf{z}$ and the predicted answer $\mathbf{y}$. We define the pseudo exemplar base $E'$ as:
\begin{equation}
E' = \{(\mathbf{x}, \mathbf{z}, \mathbf{y})|(\mathbf{x},\mathbf{y}^*)\in D, \mathbf{y}=\mathbf{y}^*\},
\end{equation}
then $E\cup E'$ can be regarded as the new exemplar base for generating diverse prompts.

\subsection{Noises in Multiple Choice Tasks}
\label{sec:noise}

In our experimental setup, StrategyQA and CommonsenseQA are more challenging than other tasks, as they use pseudo exemplars generated through ``self-teaching'' (Appendix \ref{sec:self_teaching}).

``Self-teaching'' may lead to bad exemplars, whose reasoning paths are invalid but happen to yield answers coinciding with the ground truth.
Questions in StrategyQA/CommonsenseQA are two-choice/four-choice questions, respectively.
Therefore, such noise would be more serious in StrategyQA than in CommonsenseQA.
This somehow explains why \textsc{DiVeRSe} can achieve comparable performance ($-0.8\%$) as the PaLM-based SOTA on CommonsenseQA, while it sees a $3.0\%$ performance decline to PaLM on StrategyQA, which has only two choices. In other words, it is easier for StrategyQA to yield a right answer but a misleading reasoning path. 

\section{Data Statistics}
\label{sec:data_statistics}
Table \ref{tab:data_stat} shows the reasoning benchmarks we use in this paper with examples.
We use the same test sets as \citet{wei2022chain} for GSM8K, AsDiv, MultiArith, SVAMP, SingleEq, and CommonsenseQA.

For StrategyQA, there are $2,290$ test cases (i.e., questions paired with TRUE/FALSE labels), but there is no other case that can be leveraged by \textsc{DiVeRSe} to construct diverse exemplars (as introduced in Section \ref{section:diversity}).
To address this problem, we randomly divide these $2,290$ test cases into two equal parts (denoted as $T_1$ and $T_2$).
For each \textsc{DiVeRSe} experiment of SQA, we conduct two runs: using $T_1$ to construct diverse exemplars and $T_2$ as the test set, and vice versa.
The final reported solve rate is the average solve rate of these two runs.

For CLUTRR, \citet{sinha2019clutrr} provided several versions: \textit{clean}, \textit{supporting}, \textit{irrelevant}, and \textit{disconnected}.
The \textit{clean} version is the basic dataset, while the others are the perturbed variations of it.
Our experiments are conducted on the \textit{clean} version.

\section{Our Changes to CLUTRR}
\label{sec:clutrr_detail}

In our experiments, two changes are applied to the CLUTRR benchmark:
(1) appending candidate answers to each question; (2) constructing reasoning paths based on rules. Table \ref{tab:clutrr_example} shows an example of CLUTRR data after our modification.

\paragraph{Candidate Answers.}
Besides the original questions (e.g., ``\textit{Mary, a female, took her husband who is a male, Roy, out for lunch. Ernest bought to dress for his father Roy. What kind of relative is Ernest to Mary?}''), we also provide all the candidate answers (i.e., ``\textit{The possible relationships are: sister, son, aunt, granddaughter, father, grandfather, grandmother, mother-in-law, uncle, niece, mother, brother, daughter, nephew, grandson, son-in-law, father-in-law, daughter-in-law}'') in the input sequence.
Our preliminary experiments show that, the gigantic language models cannot reach more than $50\%$ accuracy without the sequence of candidate answers.

\paragraph{Reasoning Paths.}
For each question, \citet{sinha2019clutrr} also provided a knowledge graph that formulates the relations directly mentioned in the question.
Each knowledge graph consists of several $\langle e_1, r, e_2\rangle$ triplets, which means there is a relation $r$ from $e_1$ to $e_2$.
Take the aforementioned question as an example, the knowledge graph consists of two triplets: $\langle \text{Mary}, \text{husband}, \text{Roy}\rangle$ and $\langle \text{Ernest}, \text{father}, \text{Roy}\rangle$.

For each question, we construct the reasoning path based on its knowledge graph.
We first topologically sort all triplets in the knowledge graph.
For each triplet, we convert it to a reasoning step using the template ``\textit{\{$e_2$\} is the \{$r$\} of \{$e_1$\}}''.
After that, we can get the reasoning path by concatenating these reasoning steps.
Take the aforementioned question as an example, the reasoning path is: ``\textit{Roy is the husband of Mary. Roy is the father of Ernest. Thus, Ernest is the son of Mary.}''

\begin{table}[t]
\centering
\resizebox{.99\linewidth}{!}{
\begin{tabular}{p{0.2\linewidth}<{\raggedright}p{1.0\linewidth}<{\raggedright}}
\toprule[2pt]
Variant & Input Example \\ \midrule[1.5pt]
CLUTRR for NLI (Original) &
\textbf{Input:} Story: Kelly and her mother Ernest made breakfast together. Constance and her husband Ernest wanted a child badly. Query: Kelly, Constance

\textbf{Output:} \textbf{daughter} \\
\midrule[.5pt]
CLUTRR for NLG (Ours) &
\textbf{Input:} Kelly and her mother Ernest made breakfast together. Constance and her husband Ernest wanted a child badly. What kind of relative is Kelly to Constance? \redcell{The possible relationships are: sister, son, aunt, granddaughter, father, grandfather, grandmother, mother-in-law, uncle, niece, mother, brother, daughter, nephew, grandson, son-in-law, father-in-law, daughter-in-law.} 

\textbf{Output:} \bluecell{Kelly is the daughter of Ernest. Ernest is the husband of Constance. Thus, Kelly is the daughter of Constance.} So the answer is \textbf{daughter}.\\
\midrule[2pt]
\end{tabular}
}
\caption{Examples of the original and our constructed CLUTRR data. In our experiments, we concatenated the class information (marked with \redcell{red}) after the input question as the input, and constructed reasoning paths (marked with \bluecell{blue}) via a simple text template as the output.
}
\label{tab:clutrr_example}
\end{table}

\begin{table*}[t]
\renewcommand\arraystretch{1.2}
\small
\centering
\resizebox{0.99\linewidth}{!}{
\begin{tabular}{p{0.3\textwidth}<{\raggedright}|p{0.4\textwidth}<{\raggedright}|p{0.4\textwidth}<{\raggedright}}
\toprule[2pt]
Question & Self-Consistency & \textsc{DiVeRSe} \\
\midrule[1.5pt]
\textbf{[GSM8K]} \textit{Siobhan has 2 fewer jewels than Aaron. Aaron has 5 more jewels than half of Raymond's jewels. If Raymond has 40 jewels, how many jewels does Siobhan have?} &
Aaron has 2 more jewels than half of Raymond's jewels. So Aaron has \redcell{\textbf{40 / 2 + 2 = 22}} jewels. Siobhan has 2 fewer jewels than Aaron so she has \redcell{\textbf{22 - 2 = 20}} jewels. So the answer is \redcell{\textbf{20}}.

\textbf{Answer:} \redcell{\textbf{20 (Wrong)}}\qquad
\textbf{Count:} $33$&

Raymond has 40 jewels. Half of Raymond's jewels is \bluecell{\textbf{40 / 2 = 20}}. Aaron has 5 more jewels than half of Raymond's jewels, so he has \bluecell{\textbf{20 + 5 = 25}} jewels. Siobhan has 2 fewer jewels than Aaron, so she has \bluecell{\textbf{25 - 2 = 23}} jewels. So the answer is \bluecell{\textbf{23}}.

\textbf{Answer:} \bluecell{\textbf{23 (Correct)}}\qquad
\textbf{Count:} $32$ \\ \hline

\textbf{[StrategyQA]} \textit{Could you go to New York Public Library and the Six Flags Great Escape in the same day?} &
The New York Public Library is in Manhattan. The Six Flags Great Escape is in Queensbury. Manhattan and Queensbury are too far apart to go to both in a single day. Thus, you could not go to New York Public Library and the Six Flags Great Escape in the same day. So the answer is \textbf{\redcell{no}}.

\textbf{Answer:} \textbf{\redcell{no (Wrong)}}\qquad
\textbf{Count:} $62$&

The New York Public Library is in New York City. The Six Flags Great Escape is in Queensbury, New York. Queensbury is about 3.5 hours away from New York City by car. Thus, you could go to the New York Public Library and the Six Flags Great Escape in the same day. So the answer is \textbf{\bluecell{yes}}.

\textbf{Answer:} \textbf{\bluecell{yes (Correct)}}\qquad
\textbf{Count:} $38$ \\
\bottomrule[2pt]
\end{tabular}
}
\caption{Examples of \textit{code-davinci-002} on GSM8K.
Compared to \textit{self-consistency} (majority voting), \textsc{DiVeRSe} can select the correct-but-not-most answer out of the sampled candidates, thus improving the reasoning performance.}
\label{tab:code_davinci_case}
\end{table*}

\end{document}